\documentclass{article}

\usepackage{microtype}
\usepackage{graphicx}
\usepackage{subcaption}
\usepackage{booktabs} 

\usepackage{hyperref}
\usepackage{url}
\usepackage{graphicx} 
\usepackage{booktabs} 
\usepackage{microtype}
\usepackage{adjustbox} 
\usepackage{amssymb} 
\usepackage{amsmath}
\usepackage{enumitem}
\usepackage{amsthm}
\usepackage{pifont}
\usepackage{makecell} 
\newcommand{\cmark}{\textcolor{green!60!black}{\ding{51}}} 
\newcommand{\xmark}{\textcolor{red!70!black}{\ding{55}}}   

\usepackage{wrapfig}
\usepackage{amsthm}
\usepackage{multirow}
\usepackage{tcolorbox}
\theoremstyle{definition}

\usepackage{hyperref}

\usepackage[preprint]{icml/icml2026}

\usepackage{amsmath}
\usepackage{amssymb}
\usepackage{mathtools}
\usepackage{amsthm}
\usepackage{pifont}

\usepackage[capitalize,noabbrev]{cleveref}

\theoremstyle{plain}

\theoremstyle{definition}

\theoremstyle{remark}

\usepackage[textsize=tiny]{todonotes}

\icmltitlerunning{Towards Universal Neural Likelihood Inference}

\begin{document}

\twocolumn[
  \icmltitle{Towards Universal Neural Likelihood Inference}



  \icmlsetsymbol{equal}{*}

\begin{icmlauthorlist}
  \icmlauthor{Shreyas Bhat Brahmavar}{yyy}
  \icmlauthor{Yang Li}{}
  \icmlauthor{Qiyang Liu}{yyy}
  \icmlauthor{Shashank Srivastava}{yyy}
  \icmlauthor{Junier Oliva}{yyy}
\end{icmlauthorlist}

\icmlaffiliation{yyy}{Department of Computer Science, University of North Carolina at Chapel Hill, Chapel Hill, NC, USA}

\icmlcorrespondingauthor{Shreyas Bhat Brahmavar}{shreyas@cs.unc.edu}
  \icmlkeywords{Machine Learning, ICML}

  \vskip 0.3in
]



\printAffiliationsAndNotice{}  

\begin{abstract}
We introduce universal neural likelihood inference (UNLI): enabling a single model to provide data-grounded, conditional likelihood predictions for arbitrary targets given any collection of observed features, across diverse domains and tasks. To achieve UNLI over heterogeneous tabular data, we develop the Arbitrary Set-based Permutation-Invariant Reasoning Engine (ASPIRE) model.
Our design addresses critical gaps in existing approaches to merge semantic-understanding capabilities and generalised
numerical feature reasoning within a zero-shot capable framework. Trained on over 1,400 real diverse datasets spanning various domains, ASPIRE achieves 15\% higher F1 scores and 85\% lower RMSE than existing tabular foundation models in zero-shot and few-shot settings. Lastly, this work introduces open-world active feature acquisition, where we leverage the UNLI capabilities of ASPIRE to adeptly determine next feature-values to observe to improve inference time prediction accuracies.
\end{abstract}

\section{Introduction}
In this work, we consider an important but underexplored aim—\emph{universal likelihood inference}—building a \emph{single model} imbued with an \emph{open-world} understanding of arbitrary semantic concepts \emph{and} feature dependencies toward the prediction of any specified target in the form of a normalized conditional likelihood.
For example, within medical applications, we would like \emph{a single model} to intelligently be able to make inferences for \emph{all} of the following tasks:
\begin{itemize}[label={$\bullet$}, topsep=0pt, noitemsep, leftmargin=*]
  \item Provide primary-care risk assessment for low-resource rural patients, estimating a given patient’s 5-year colorectal cancer risk using only the subset of information from a routine visit, age, sex, family history of cancer, haemoglobin, and self-reported bowel symptoms (e.g., while imaging, prior colonoscopy results, and genetic markers remain unobserved).
  \item Provide population-level health surveillance and planning in urban settings, estimating a conditional pdf of hospitalization rates for respiratory illness in a given region using only observed signals such as air quality, vaccination coverage, mobility patterns, and weather indicators.
  \item Provide clinical evaluation to infer a conditional pdf for a given patient's missing lab measurement (e.g., haemoglobin)  conditioned on whatever evidence is already present, including demographics, fatigue severity, medication history, and prior diagnoses.\looseness-1
\end{itemize}

More generally, our goal is a model that produces quantitative probabilistic predictions (explicit likelihoods over target domains) across a broad, open-world family of tasks spanning heterogeneous contexts, feature subsets, and prediction targets. Such a capability enables data-grounded statistical inference \emph{even in unanticipated settings}, including \emph{regimes where labeled data are scarce or entirely unavailable}.

\begin{figure}[ht]
    \centering
    \includegraphics[width=.75\linewidth]{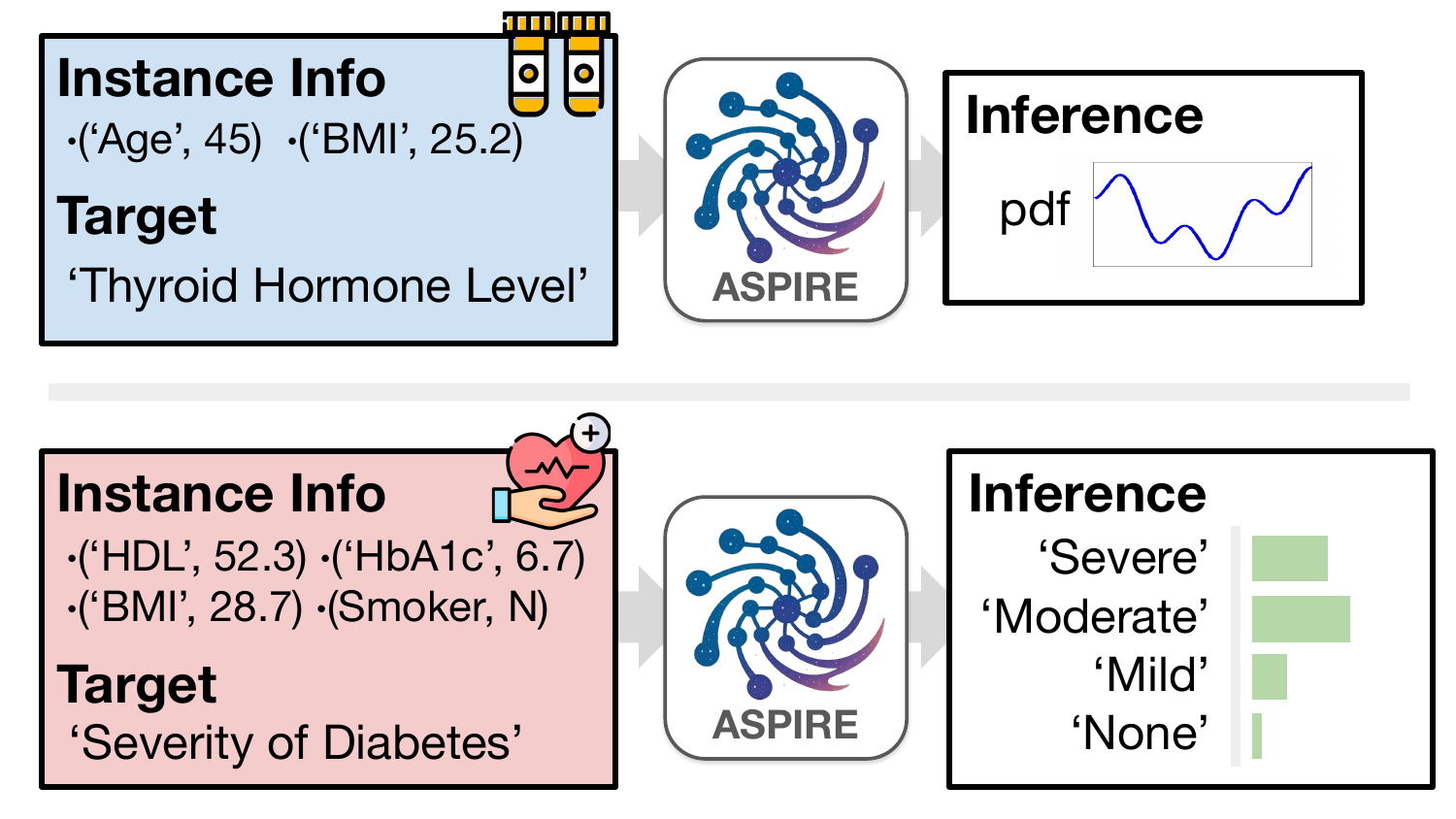}
    \caption{Illustration of UNLI capabilities, where a single ASPIRE model is able to output a likelihood in the domain of the specified target over a wide universe of conditioning information.}
    \label{fig:cartoon}
\end{figure}


\begin{table*}[t]
\centering
\caption{Representative tabular foundation models across relevant UNLI capabilities (\ding{81} denotes partial capabilities, expounded in text).}
\label{tab:tabular_fm_comparison}
\footnotesize
\setlength{\tabcolsep}{4pt}
\begin{tabular}{l*{8}{c}}
\toprule
Method & 
\makecell{Arbitrary\\Conditioning} & 
\makecell{PIV\\Features} & 
\makecell{PIV\\Shots} & 
\makecell{Semantic\\Context} & 
\makecell{Zero-shot\\Inference} &
\makecell{In-Context\\Learning} \\
\midrule
CARTE~\cite{kim2024carte}              & \xmark & \cmark & \xmark & \cmark & \xmark & \xmark \\
TP-BERTa~\cite{yan2024making}          & \xmark & \cmark & \xmark & \cmark & \xmark & \xmark \\
CM2~\cite{ye2024towards}               & \cmark & \cmark & \xmark & \cmark & \cmark & \xmark \\
TabLLM~\cite{hegselmann2023tabllm}     & \cmark & \xmark & \xmark & \cmark & \cmark & \cmark \\
GTL~\cite{wen2024supervised}           & \cmark & \xmark & \xmark & \cmark & \cmark & \cmark \\
TabPFNv2.5~\cite{hollmann2025accurate} & \cmark & \ding{81} & \cmark & \xmark & \xmark & \cmark \\
TabICL~\cite{qu2025tabicl}             & \cmark & \xmark & \cmark & \xmark & \xmark & \cmark \\
LimiX~\cite{zhang2025limix}            & \cmark & \xmark & \cmark & \xmark & \xmark & \cmark \\
\midrule
\textbf{ASPIRE}                         & \cmark & \cmark & \cmark & \cmark & \cmark & \cmark\\
\bottomrule
\end{tabular}
\end{table*}

\textbf{Aim}\quad 
Unlike many existing cross-tabular approaches \cite{yan2024making, ye2024towards}, which focus on cross-dataset learning to provide more efficacious transfer learning for test time \emph{datasets} (where hundreds to over many thousands of training instances are observed), our focus is \textbf{on-the-spot} conditional likelihood inference to predict an arbitrary given target with only a particular query instance (\emph{zero-shot}), or with an \emph{optional} additional context of a very small ($\leq 5$) collection of labeled examples (few-shot)\footnote{We additionally also show SOTA performance when predicting in many-shot settings.}. We coin this focus \textbf{universal neural likelihood inference} (UNLI).

\textbf{Approach}\quad 
We formalize universal neural inference over heterogeneous structured data and propose a principled solution for \emph{arbitrary feature sets and targets}: the Arbitrary Set-based Permutation-Invariant Reasoning Engine (ASPIRE). ASPIRE frames inference as a set-based reasoning problem, modeling both instance features and support examples as unordered sets, thereby ensuring permutation invariance while enabling flexible reasoning over arbitrary feature combinations. Its semantic grounding mechanism leverages natural-language feature descriptions, dataset metadata, and in-context examples to learn meaningful dependencies across datasets. By conditioning representations on semantic information rather than feature positions, ASPIRE aligns shared concepts across heterogeneous schemas and generalizes to new datasets without additional training. Importantly, ASPIRE’s likelihood-based interface naturally supports downstream tasks such as Active Feature Acquisition (AFA) \cite{saar2009active, li2021active}, where principled uncertainty estimates and arbitrary conditioning are essential for adaptively selecting informative features at inference time.\looseness-1
\textbf{Contributions}\quad
In summary, ASPIRE is a novel approach that bridges several critical gaps in current day cross tabular methodology. ASPIRE is cohesive, unifying approach that: 1) is able to attain the rich semantic understanding of LLM meta-data/schema-based methods \cite{wen2024supervised,yan2024making,ye2024towards} along with the generalized numerical feature reasoning of tabular foundation models \cite{hollmann2022tabpfn,qu2025tabicl,zhang2025limix}; 2) achieves the desired invariances between \emph{both} query instance features and in context labeled shots; 3) abstracts multiple permutation equivariant modules that enable the late fusion of input information for inference; 4) frames training with UNLI, a self-supervised, multi-domain likelihood loss; 5) is able to perform zero, few, and many shot test-time inference within the same model; 6) achieves state-of-the-art performance across multiple settings (zero, few, and many shots) in both classification and regression tasks; 7) represents the first use of generalized inference toward open-world active feature acquisition \cite{ma2018eddi,li2021active,li2024distribution}, where the universality of our ASPIRE model is especially adept to discover inference time subsets of informative features to observe for improved prediction.\looseness-1

\section{Background and Related Works}\label{sec:background}




Universal neural likelihood inference (UNLI) requires a model to provide normalized conditional probability distributions for arbitrary targets given any observed feature subset, generalizing zero-shot across heterogeneous datasets. This demands five critical capabilities: (1) \textbf{universal arbitrary conditioning}—predicting any variable given any other subset (without retraining); (2) \textbf{permutation invariance}—ensuring predictions are independent of feature and instance orderings; (3) \textbf{semantic grounding}—aligning features across datasets through natural language understanding (4) \textbf{zero-shot inference}—reasoning solely given a query instance; and (5) \textbf{in-context learning}—additionally conditioning on a small number of labeled shots when provided. Although existing work addresses these capabilities in isolation, no prior method simultaneously achieves all requirements for UNLI (see Tab.~\ref{tab:tabular_fm_comparison}); our work (ASPIRE) can be seen as a novel approach to address these challenges and bridge the gaps for UNLI in the current methodology.\looseness-1

\textbf{Tabular Foundation Models} \quad
Recent tabular foundation models have made significant progress toward cross-dataset learning, yet each approach addresses only a subset of UNLI requirements. 
Early work on cross-table learning focused on developing transferable representations that generalize across datasets. XTab~\cite{zhu2023xtab}, CARTE~\cite{kim2024carte}, TaBERT~\cite{yin2020tabert}, TP-BERTa~\cite{yan2024making}, and CM2~\cite{ye2024towards} learn shared backbones that can be fine-tuned for new tasks. Recent serialization-based methods like TabLLM~\cite{hegselmann2023tabllm} and GTL~\cite{wen2024supervised} convert rows to natural language sequences (e.g., ``[Age] is 32, [Gender] is male''), grounding structured data in the semantic knowledge of pre-trained LLMs. FeatLLM~\cite{han2024largelanguagemodelsautomatically} uses an LLM as a feature engineer, generating rule-based derived features from task/feature descriptions and few-shot examples, after which a lightweight predictor is trained on these engineered features. While these methods achieve semantic grounding and cross-dataset transfer, they rely on sequence models with positional encodings, making predictions sensitive to feature ordering—a critical violation of permutation invariance.\looseness-1

A paradigm shift occurred with TabPFN~\cite{hollmann2022tabpfn}, which interprets in-context learning (ICL) from a Bayesian perspective as approximate posterior predictive inference. Pretrained on extensive synthetic datasets, TabPFN predicts directly from examples provided in the prompt, treating support examples as unordered sets to achieve instance-level permutation invariance. TabPFNv2~\cite{hollmann2025accurate} substantially improved both prediction performance and scalability. 
Several variants have emerged: TabForestPFN~\cite{den2024fine} incorporates tree-based synthetic datasets; TabDPT~\cite{ma2025tabdpt} uses real-world pretraining data; LimiX \cite{zhang2025limix} represents the most comprehensive prior effort, achieving arbitrary conditioning through joint generative modeling of feature values and missingness; TabICL~\cite{qu2025tabicl} achieves comparable performance with improved computational efficiency through column-then-row attention mechanisms.\looseness-1
\textbf{Arbitrary Conditional Distribution}\quad
Arbitrary conditioning—predicting any subset of variables given any other subset—is a core primitive for probabilistic inference. Prior work achieves this \emph{within a fixed feature universe} within a single dataset. The Universal Marginalizer~\citep{douglas2017universal} predicts missing variables from observed subsets (often with factorization assumptions), while VAEAC~\citep{ivanov2018variational}, ACFlow~\citep{li2020acflow}, and ACE~\citep{strauss2021arbitrary} provide increasingly expressive conditional models with learned dependencies; Neural Conditioner~\citep{belghazi2019learning} supports arbitrary conditioning but does not yield tractable likelihoods. Crucially, these methods generalize over \emph{missingness patterns} in a closed-world \emph{single dataset} setting, rather than across datasets with heterogeneous schemas. ASPIRE extends arbitrary conditioning to the \emph{cross-dataset} regime required by UNLI.


\textbf{Permutation Invariant Set Modeling} \quad 
Models performing UNLI should not be sensitive to the ordering \emph{of features} nor \emph{of labeled shots}. That is, if the same set of features or instances is presented, the input information is equivalent, and thus inference should not depend on any inter order in which elements are kept.
Formally, a function $f: \mathcal{X}^n \rightarrow \mathcal{Y}$ is \textbf{permutation invariant} (PIV) if $f(\pi(\mathbf{x})) = f(\mathbf{x})$ for any permutation $\pi$, and \textbf{permutation equivariant} if $f(\pi(\mathbf{x})) = \pi(f(\mathbf{x}))$ when the output is also a set. \looseness-1

Two primary frameworks enable permutation-invariant modeling: \textbf{DeepSets}~\citep{zaheer2017deep} uses pooling-based aggregation (e.g., sum or mean pooling) over element-wise transformations, while \textbf{Set Transformer}~\citep{lee2019set} uses attention-based aggregation through self-attention mechanisms to enable richer element interactions while maintaining permutation equivariance. Further developments include generalized aggregation functions~\citep{kimura2024permutation}, subset-invariant regularization~\citep{cohen2020regularizing}, and theoretical analyses of capacity-size trade-offs~\citep{wang2023polynomial}.


ASPIRE builds on these principles to achieve \emph{dual} \textbf{feature-level} and \textbf{instance-level} \emph{permutation invariance} through hierarchical set-based processing.
TabPFNv2 variants~\citep{hollmann2025accurate} interleave row-wise and column-wise attention to address permutation invariance; however, \citeauthor{hollmann2025accurate}~note, ``To allow our model to differentiate features more easily that have the same statistics, for example, two features that have the same entries just in different orders, we use random feature embeddings that we add to all embeddings before the first layer." These random feature embeddings break a strict invariance over the order of features (as seed and feature order now affect inference) \cite{ye2025closer, qu2025tabicl} (denoted as \ding{81} in Tab.~\ref{tab:tabular_fm_comparison}).
ASPIRE uniquely combines dual permutation invariance plus semantic grounding through natural language descriptions, enabling consistent UNLI with zero-shot generalization regardless of internal information ordering.

\section{Method}
We develop our UNLI loss and inference framework  ($\S$\ref{sec:UNLI}), to train ASPIRE's architecture  ($\S$\ref{sec:arch}). \looseness-1

\subsection{Universal Neural Likelihood Framework}
\label{sec:UNLI}
We propose and formalize the framework of \textbf{universal neural likelihood inference}: learning a single model that can estimate the likelihood for \emph{any target} (a pmf in discrete domains or a pdf in continuous domains) given a corresponding \emph{arbitrary} feature subset. We note that, unlike previous approaches that tokenize or discretize real-valued targets \cite{hegselmann2023tabllm,wen2024supervised,yan2024making} or only provide point estimates \cite{ye2024towards}, we consider target likelihoods in their native domains, enabling better uncertainty characterization, multiple mode estimation, and better information gain inference.

\textbf{Dataset and instance representation} \quad
Let $\mathcal{D} = \{\mathcal{D}_k\}_{k=1}^{K}$ denote a collection of datasets, where each dataset $\mathcal{D}_k = \{e_n\}_{n=1}^{N_k}$ contains $N_k$ i.i.d. examples. Each example $e_n$ is represented as a set of $M_k$ feature-value pairs: $e_n = \{(f_m, v_m)\}_{m=1}^{M_k}$, where $f \in \mathcal{F}$ and $\mathcal{F}$ represents the universe of all potential features and their metadata (e.g., textual descriptions, data types) across datasets. This set-based representation naturally accommodates heterogeneous real-world datasets where feature sets, types, and semantics vary significantly. A core principle of our approach is processing each instance as an \emph{unordered set} of feature-value pairs whilst ingesting textual context, which is in contrast to tokenization/serialization approaches~\citep{hegselmann2023tabllm,wen2024supervised} that impose specific orderings and violate permutation invariance.

\textbf{Universal Arbitrary Conditioning} \quad
Previous arbitrary conditioning methods~\citep{ivanov2018variational,li2020acflow,strauss2021arbitrary} operate within a single dataset where features are restricted to those co-occurring during training. We extend this to \textbf{universal arbitrary conditioning} across datasets (Eq.~\ref{eq:few-shot}), where any subset of features from the broad universe $\mathcal{F}$ may be observed for a given instance, and the task is to predict any unobserved target.
%
%
To achieve UNLI across diverse datasets, we must train our model to handle three critical challenges that extend beyond single-dataset arbitrary conditioning: (1) \emph{semantic heterogeneity}: features with similar meanings have different names and representations across datasets (e.g., ``patient age'' vs. ``age (years)''); (2) \emph{zero-shot generalization}: the model must perform inference on entirely unseen datasets without retraining; (3) \emph{few-shot adaptation}: when available, a small number of labeled examples from the target dataset should improve predictions. 
To address these challenges, we augment the typical arbitrary conditioning likelihood by conditioning on:
\begin{itemize}[leftmargin=*,topsep=0pt, noitemsep]
    \item \textbf{Dataset context $c$}: Text description of the dataset and its domain (e.g., ``medical records from cardiology patients''), embedded as tokens with positional encodings. This provides high-level semantic context about the data distribution and helps align concepts across domains.\looseness-1
    \item \textbf{Target description $f_t$}: similarly specifies what to predict—the identities and metadata of unobserved features, enabling the model to understand the prediction task.
    \item \textbf{Observed \emph{set} of feature-value pairs $\{(f_m, v_m)\}_{m \in o}$}: provide the evidence for prediction—both the feature identities (descriptions/meta-data) and their values for the query instance.
    \item (\emph{Optional}) \textbf{Support set $S$}: A small set of labeled examples $S = \{e_s\}_{s=1}^{|S|}$ from the same dataset, where each $e_s = \{(f_{m}, v_{m}^{(s)})\}_{m=1}^{M} \cup \{(f_t, v_{t}^{(s)})\}$ is fully observed. 
    When $S = \emptyset$, the model performs zero-shot inference.
\end{itemize}

\textbf{Universal Likelihood Maximization} \quad 
During training, we sample from a \emph{task distribution} $\mathcal{T}$ that simulates the diversity of scenarios our model will encounter at test time: select a dataset uniformly from $\mathcal{D}$, randomly choose observed features $o$ and target feature $t$, sample a query instance, and optionally sample a support set $S$ of random cardinality (0-5 examples). Our model parameters, $\gamma$, are trained to minimize the expected negative log-likelihood:
\begin{equation}\label{eq:few-shot}
\mathbb{E}_{\mathcal{T}} \Big[ -\log p_\gamma \big(v_t \mid \{(f_m, v_m)\}_{m \in o}, f_t, S, c \big) \Big].
\end{equation}

\begin{figure}[t]
    \centering
    \includegraphics[width=\linewidth]{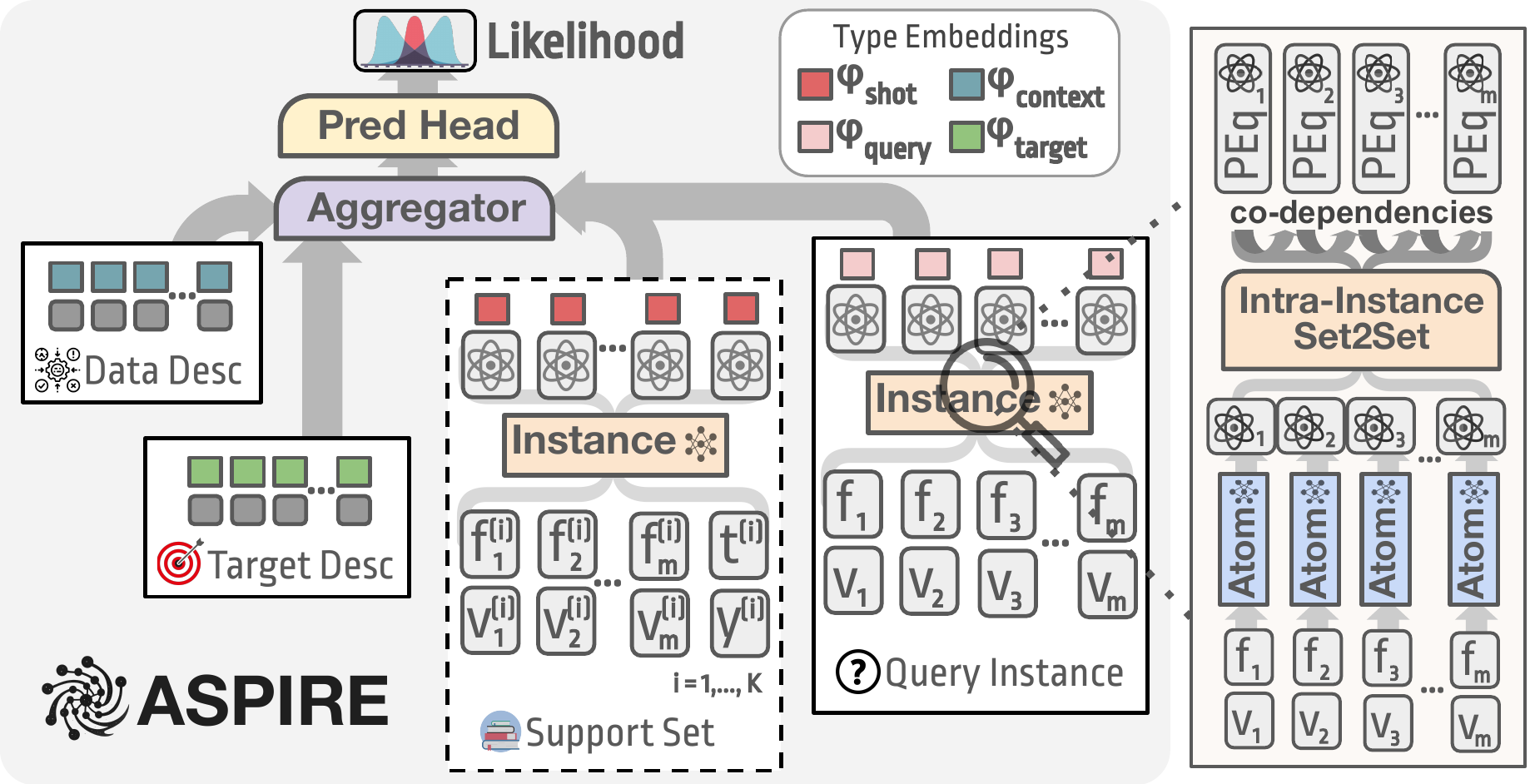}
    \caption{ASPIRE architecture. The model processes query instances, optional support sets, and dataset context through set-based mappings, maintaining permutation invariance at both feature and instance levels for UNLI.}
    \label{fig:uni}
    
\end{figure}
\subsection{Architecture}
\label{sec:arch}

We next describe ASPIRE, our neural architecture designed to satisfy the UNLI interface and efficiently optimize the Universal Likelihood Maximization objective.
A founding principle behind ASPIRE is the permutation invariant processing of \emph{sets} of information. We develop this through a hierarchal, modular design that abstracts various units of information and their fusion (see Fig.~\ref{fig:uni}):
\begin{enumerate}[leftmargin=*,topsep=0pt, noitemsep]
\item \textbf{Atom Processing} (\S\ref{sec:atoms}) converts feature-value pairs into informational "atoms" where values are embedded conditionally on their feature semantics, ensuring that ``32'' is processed differently for ``age'' versus ``BMI''.
\item \textbf{Intra-Instance Representation} (\S\ref{sec:sets}) treats each instance as a set of atoms, and learns a permutation equivariant \emph{set-to-set} mapping to output a corresponding set of embedding elements that are each influenced (depend on) the co-occurring atoms within the instance; i.e., $M$ separate feature-value atoms are mapped to $M$ output embeddings that are \emph{each} reflective of co-occurring feature-values within the instance, allowing for late-fusion.
\item \textbf{Inter-Instance Aggregation} (\S\ref{sec:aggregation}) fuses intra-instance-aware atom representations from query and support instances along with natural language tokens from dataset context. Each representation is tagged with learnable type embeddings to distinguish its role: observed query features provide evidence, target features require prediction, support examples provide task-specific context, and dataset descriptions provide semantic grounding. Tagged representations are aggregated through Set Transformers to enable cross-attention while maintaining permutation invariance, allowing the model to integrate structural patterns from examples with semantic knowledge from descriptions.
\item \textbf{Prediction Heads} (\S\ref{sec:heads}) process the aggregated representation to output target-specific likelihoods: pdfs continuous targets and categorical pmfs for discrete targets.
\end{enumerate}

\subsubsection{Feature-Value Atom Processing}\label{sec:atoms}
Each instance consists of feature-value pairs (for each considered feature) that we term ``atoms," the fundamental units of information in ASPIRE. A key challenge is achieving contextual awareness: identical values must be embedded differently depending on their feature context. For example, ``32'' should be processed differently for ``age'' versus ``BMI''.
We address this through feature-conditioned value embedding. \looseness-1

\textbf{Values} are encoded according to their data type:
\begin{equation}
\nu(v_m | f_m) = \begin{cases}
\text{LE}(v_m) & \text{if $f_m$ categorical} \\
\text{Fourier}(v_m) & \text{if $f_m$ continuous} \\
\mathbf{e}_{\emptyset} & \text{if $v_m$ missing}
\end{cases}
\end{equation}
where $\text{LE}(v_m)$ is a language embedding of the specific categorical value, e.g., ``hispanic'', ``male'', (for example, using BERT~\citep{devlin2019bert}), Fourier encoding~\citep{zhou2025fone} provides single-token representations for numerical values, and $\mathbf{e}_{\emptyset}$ is a learnable missing value embedding shared across datasets. 

\textbf{Features} (descriptions) are mapped into a shared representation space using metadata-aware embeddings:
\begin{equation}
\phi(f_m) = E_{\text{desc}} + E_{\text{dtype}} + \mathbb{I}(\text{dtype} = \text{Cat.}) \cdot E_{\text{choices}},
\label{eq:sem}
\end{equation}
where $E_{\text{desc}}$ encodes text descriptions (e.g., using BERT~\citep{devlin2019bert}), $E_{\text{dtype}}$ provides learnable type embeddings (e.g., numerical/categorical), and $E_{\text{choices}}$ aggregates categorical possible value representations when applicable.\looseness-1 




\textbf{Atom} embeddings fuse the value encoding with semantic feature information:\looseness-1
\begin{equation}
\psi(f_m, v_m) = \text{AtomMLP} \big( \phi(f_m), \nu(v_m | f_m) \big)
\label{eq:psi}
\end{equation}
where AtomMLP learns to modulate value representations based on feature semantics $\phi(f_m)$ (Eq.~\ref{eq:sem}), ensuring that identical values receive different contextual encodings depending on their feature context.


\subsubsection{Intra-Instance Representation}\label{sec:sets}
ASPIRE treats each instance as an unordered set of atoms $e_n = \{(f_m, v_m)\}_{m=1}^{M}$, where $v_m$ might be missing for unobserved features. Rather than compressing this set into a single vector, we propose to employ permutation-equivariant (PEq) \emph{set-to-set} mappings that maintain set structure while enabling feature interactions (co-dependencies based on instance co-occurring feature-value pairs).

As illustrated in Figure~\ref{fig:uni}, the set-to-set mapping processes atom embeddings to yield multiple output vectors (one per atom) that capture intra-dependencies between co-occurring features. This allows feature representations to become instance-aware without bottlenecking through a single compressed vector, enabling \emph{late-fusion} of instance information.

We apply stacked Set Transformer layers \citep{lee2019set} to the atom embeddings, $\rho(e_n) = \{\rho_1(e_n), \ldots, \rho_M(e_n)\}$:
\begin{equation}\label{eq:rho}
\rho(e_n) = \text{IntraSet2Set}\big( \big\{ \psi(f_m, v_m) \big\}_{m=1}^{M} \big).
\end{equation}
The Set Transformer architecture is well-suited for this task because both its attention mechanism and feed-forward layers are inherently permutation-equivariant. The final atom embeddings incorporate instance-level contextual information, allowing each feature-value pair to be aware of other features within the same instance.

In few-shot learning scenarios, the model additionally receives a variable number of labeled examples (support set) $S = \{e_s\}_{s=1}^{|S|}$ from the same dataset. Each support example $e_s$ consists of observed feature-value pairs and is processed through the same instance embedding module, ensuring unified representation across query and support instances:
\begin{equation}
    \Lambda(S) = \big\{ \rho(e_s) \big\}_{s=1}^{|S|}.
\end{equation}
Our set-based approach naturally provides two critical advantages: (1) \emph{permutation invariance}—feature ordering does not affect representations when utilizing our equivariant mappings composed with invariant aggregation, and (2) \emph{flexible conditioning}—any subset of features and support set can be jointly observed or predicted without requiring architectural modifications.

\subsubsection{Inter-Instance Information Aggregation}\label{sec:aggregation}
We now describe how ASPIRE aggregates these representations across multiple sources while maintaining permutation invariance. 
The model combines three types of information: 
\begin{itemize}[leftmargin=*,topsep=0pt, noitemsep]
    \item \textbf{Query instance}, $e_q = \{(f_m, v_m)\}_{m \in o} \cup \{(f_t, \varnothing)\}$, where the corresponding  output to the missing query target, $\{(f_t, \varnothing)\}$, is paired with a \texttt{CLS} token (Eq.~\ref{eq:aggtokens}) for aggregation, see below.
    
    \item \textbf{Support set}, $S = \{e_s\}_{s=1}^{K}$, where each $e_s$ is fully observed and sampled from the same dataset.

    \item \textbf{Dataset and target description},
    $\mathbf{C}_{\text{data}}$ and $\mathbf{C}_{\text{targ}}$, containing natural language descriptions of the dataset and prediction target embedded as a token sequence.

    
\end{itemize}
To enable cross-source aggregation, we tag each atom representation with learnable type embeddings ($\varphi_{\text{query}}, \varphi_{\text{target}}$, $\varphi_{\text{shot}}$, $\varphi_{\text{context}}$, and $\texttt{CLS}$) that distinguish their roles:\looseness-1
\begin{align}
\mathcal{T}_{\text{query}} &= \{[\rho_m(e_q), \varphi_{\text{query}}]\}_{m \in o} \cup \{ [\rho_t(e_q), \texttt{CLS}] \} \nonumber\\
\mathcal{T}_{\text{support}} &= \textstyle \bigcup_{s=1}^{|S|} \{[\rho_m(e_s), \varphi_{\text{shot}}]\}_{m=1}^{M} \label{eq:aggtokens} \\
\mathcal{T}_{\text{context}} &= \{[c_t, \varphi_{\text{data}}] \!:\!c_t \!\in\! \mathbf{C}_{\text{data}}\} \cup \{[c_t, \varphi_{\text{target}}] \!:\! c_t \!\in\! \mathbf{C}_{\text{targ}}\} \nonumber
\end{align}
where $[\cdot, \cdot]$ denotes concatenation. These type embeddings enable the model to distinguish between observed query features (evidence), target features (to predict), support examples (task-specific context), and dataset descriptions (semantic grounding).

Representations are aggregated through stacked Set Transformer layers \citep{lee2019set}:
\begin{equation}
r_t = \text{InterAggr}(\mathcal{T}_{\text{query}} \cup \mathcal{T}_{\text{support}} \cup \mathcal{T}_{\text{context}}),
\end{equation}
where we use the final output embedding corresponding to the \texttt{CLS} token (Eq.~\ref{eq:aggtokens}) as a global, \emph{permutation-invariant} summary representation, which is passed to the appropriate prediction head (see below).
\subsubsection{Prediction Heads}\label{sec:heads}
We propose to forward $r_t$ to a \emph{Categorical Head} or \emph{Continuous Head}, for categorical or scalar targets, respectively, to produce an appropriate likelihood (pmf or pdf) for inference and MLE (Eq.~\ref{eq:few-shot}).
The \textbf{Categorical Head} computes logits via dot-products between logits of $\mathbf{r}_t$ and language embeddings (e.g., using BERT~\citep{devlin2019bert}) of possible category values. For a categorical feature $f_t$ with possible values $\{\alpha_1, \ldots, \alpha_L\}$:
$
p(v_t = \alpha_\ell \mid r_t) = \frac{\exp(\mathbf{r}_t^\top \text{LE}(\alpha_\ell))}{\sum_{\ell'=1}^{L} \exp(\mathbf{r}_t^\top \text{LE}(\alpha_{\ell'}))},
$
where $\text{LE}(\alpha_\ell)$ is the same embedding used in §\ref{sec:atoms} for encoding observed categorical values. 
From the aggregated representation $r_t$, 
and for the target feature $f_t$. 
The \textbf{Continuous Head} models pdfs as mixtures of Gaussians:
$
p(v_t \mid r_t) = \sum_{i=1}^{I} \omega_{i} \mathcal{N}(v_t; \mu_i, \sigma_i^2),
$
where mixture parameters $\langle\omega_i, \mu_i, \sigma_i\rangle_{i=1}^I = \mathrm{ContHead}_\theta(\mathbf{r}_t)$ are the outputs of learned MLP layers $\mathrm{ContHead}$.



\subsection{Open-World Active Feature Acquisition}\label{sec:afa_method}
We present open-world active feature acquisition (AFA) as an application of UNLI, illustrating how a universal likelihood interface enables downstream decision-making without retraining.
In many real-world settings, information (feature values) come at a cost (of time, money, etc.) and often instances are only partially known, with the ability to sequentially acquire more feature values at inference time for better predictions.
Active Feature Acquisition (AFA) \citep{saar2009active} addresses this by adaptively selecting which features to acquire based on their expected informativeness for the target. However, prior work has focused squarely on \emph{closed-world} AFA, where the model only needs to consider acquisition within features of \emph{single} dataset. Inspired by universal inference capabilities of ASPIRE, here we propose a novel \emph{open-world} variant of AFA that reasons over a broad universe of features, targets, and contexts.\looseness-1


\textbf{Open-World EIG-Based Policy}\quad
We propose an \emph{expected information gain} (EIG) based policy \citep{li2021active, ma2018eddi, gadgilestimating} that can operate over \emph{broad} settings, utilizing ASPIRE as a backbone to estimate (EIG). At each step, the policy considers a partially known instance $e_o$, that can acquire an additional feature, $f$, from a broad universe $f \in \mathcal{U}$, toward the improved prediction of an arbitrary target, $v_t$.
For a candidate feature $f \in \mathcal{U}$, 
$\text{EIG}(f; e_o) =$:
\begin{equation}\label{eq:eig}
\mathcal{H}\left(p(v_t \mid e_o)\right) - \mathbb{E}_{v_f \mid x_o} \left[\mathcal{H}\left(p(v_t \mid e_o \cup \{(f, v_f)\})\right)\right],
\end{equation}
where $\mathcal{H}(\cdot)$ denotes predictive entropy, and $p(v \mid e)$ denotes the corresponding conditional likelihoods. In practice, the ground truth likelihoods $p(v_t \mid e_o)$, $p(v_f \mid e_o)$, and $p(v_t \mid e_o \cup \{(f, v_f)\})$ are unknown; however, these are exactly the likelihoods that may be estimated (along with any additional available context) using our ASPIRE model (Eq.~\ref{eq:few-shot}). It is straightforward to use these ASPIRE estimating likelihoods to estimate Eq.~\ref{eq:eig} via Monte Carlo sampling.
Note that this approach enables one to utilize a \emph{single} ASPIRE model to leverage UNLI to perform active feature acquisition in diverse scenarios, with a model that was trained with broad, multi-dataset patterns (e.g., for rapid few-show capabilities to acquire information in novel settings).


\section{Experiments}\label{experiments}
We train ASPIRE on diverse set of $\underline{\mathbf{1400}}$ tabular datasets from OpenTabs~\citep{ye2024towards}
, each with an extracted natural language dataset and feature descriptions. We evaluate on the same 20 held-out tasks  as \citet{ye2024towards}
from UCI, OpenML, and Kaggle spanning healthcare, finance, and science domains (10-500 features, 1K-1M samples).
ASPIRE uses BERT~\citep{devlin2019bert} for natural language encoding and an 8-layer SetTransformer backbone for set-based representations. We report classification performance using F1 and regression performance using RMSE, on each dataset averaged over five random seeds. During training, we use LLM-generated natural-language augmentations of dataset- and feature-level descriptions to increase linguistic diversity and improve robustness. To ensure a fair and uniform comparison, we align the input representations across methods so that all models observe the same information as ASPIRE wherever supported by the baseline model. When supported by the model architecture and training protocol, we train baselines from scratch with hyperparameter tuning. For methods that rely on large pretrained components or do not expose a full training pipeline, we follow the authors’ recommended pretrained checkpoints and evaluation procedures. Upon publication, we will make our code and data public. Please see Sec.~\ref{sec:background} for baseline citations and  App.~\ref{sec:appx_additional} for more details.

\subsection{Few-Shot Learning}
In line with the generalizability goals of UNLI, we first study few-shot inference, where models must make predictions on held-out datasets—primarily involving previously unseen features and targets—given only a minuscule support set of at most five labeled examples ($|S|\leq 5$). We evaluate predictive performance using standard classification and regression metrics (e.g., F1 and RMSE), and further examine models’ ability to reason about feature information in a few-shot, open-world Active Feature Acquisition (AFA) setting.\looseness-1

\subsubsection{Few-Shot Prediction}

\begin{figure}[t]
    \vspace{-0pt}
    \centering
    \includegraphics[width=.8\linewidth]{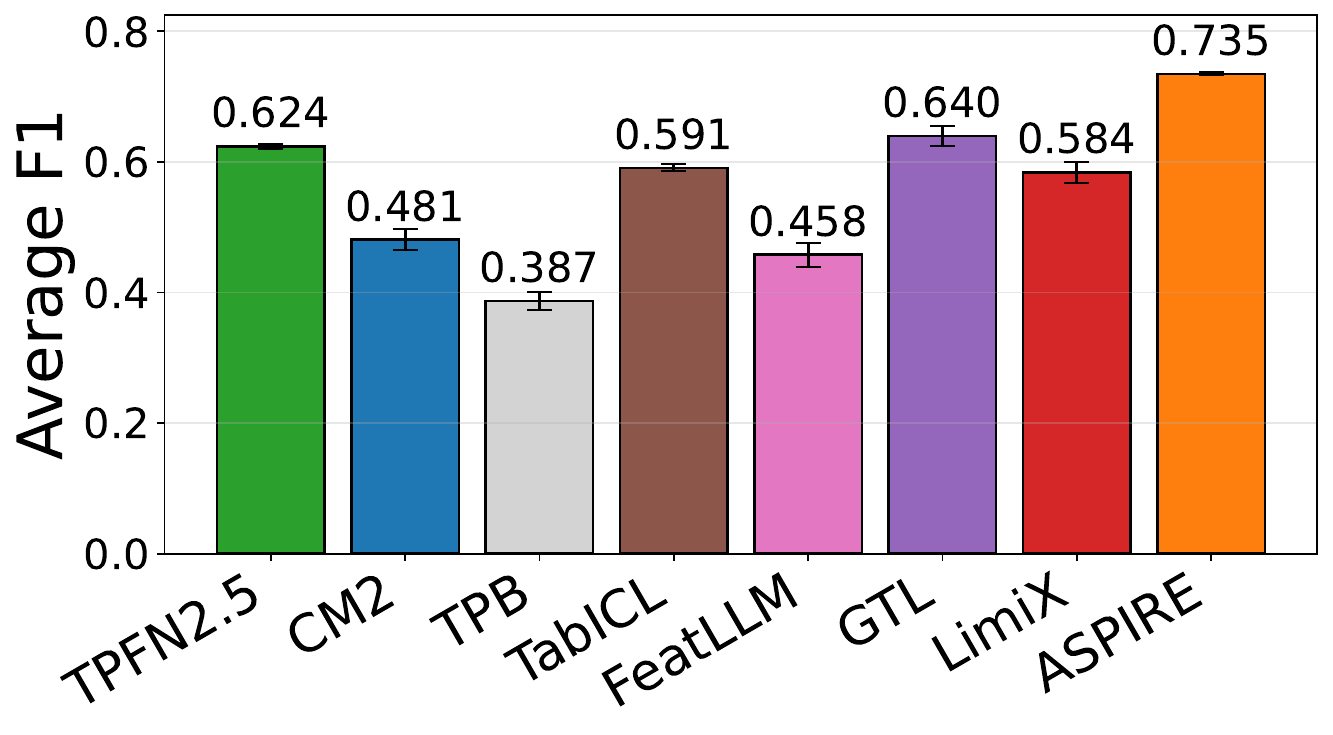}
    \caption{Five-shot F1 scores ($\uparrow$) averaged over 15 heldout classification tasks.}
    \label{fig:fewshot_class}


\end{figure}

We first evaluate models’ ability to perform few-shot reasoning on unseen classification tasks using 15 held-out datasets, following the protocol of \citet{ye2024towards}. Each baseline is evaluated using its recommended few-shot inference procedure (e.g., in-context conditioning when applicable; see App.~\ref{sec:appx_additional} for details).
Figure~\ref{fig:fewshot_class} reports average F1 scores across datasets (full results in App.~\ref{sec:appx_additional}). ASPIRE consistently outperforms all baselines, including methods such as TabPFN-v2.5 and TabICL that rely on extensive synthetic-data pretraining. Notably, other semantically grounded approaches (e.g., GTL and CM2) also fall short of ASPIRE, highlighting the advantage conferred by our UNLI formulation and ASPIRE’s set-based, likelihood-driven design.\looseness-1

\begin{figure}[ht]
\centering
\vspace{-0pt}
\includegraphics[width=.7\linewidth]{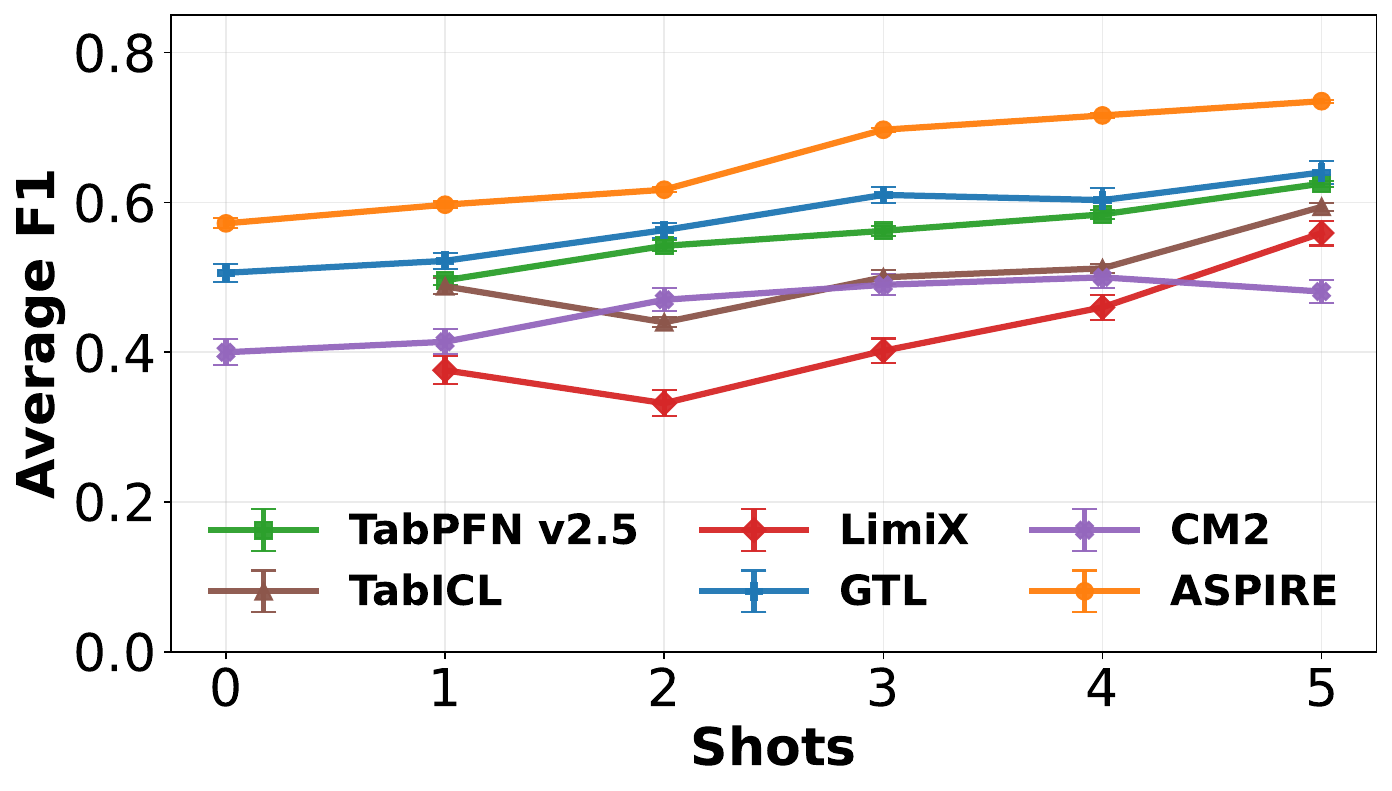}
\caption{Few-shot classification performance (Average F1) vs.\ number of shots.}
\label{fig:few_shot_f1_plot}
\end{figure}
We next examine how effectively models leverage varying amounts of contextual supervision by increasing the support set size (number of shots) and measuring the resulting classification performance. Figure~\ref{fig:few_shot_f1_plot} reports average F1 scores from zero-shot through five-shot settings across the held-out datasets, revealing several key trends. First, ASPIRE outperforms other zero-shot–capable methods, GTL and CM2, when no labeled examples are available. Notably, ASPIRE’s zero-shot performance is already competitive with the five-shot performance of several baselines, indicating that the UNLI framework enables meaningful open-world inference using only a query instance and semantic grounding. Finally, ASPIRE exhibits the most consistent performance gains as additional shots are provided, suggesting that it effectively exploits conditioning information as the available context grows.\looseness-1

\begin{wrapfigure}{r}{0.45\linewidth}
    \vspace{-10pt}
    \centering
    \begin{subfigure}{\linewidth}
    \centering
    \includegraphics[width=.85\linewidth]{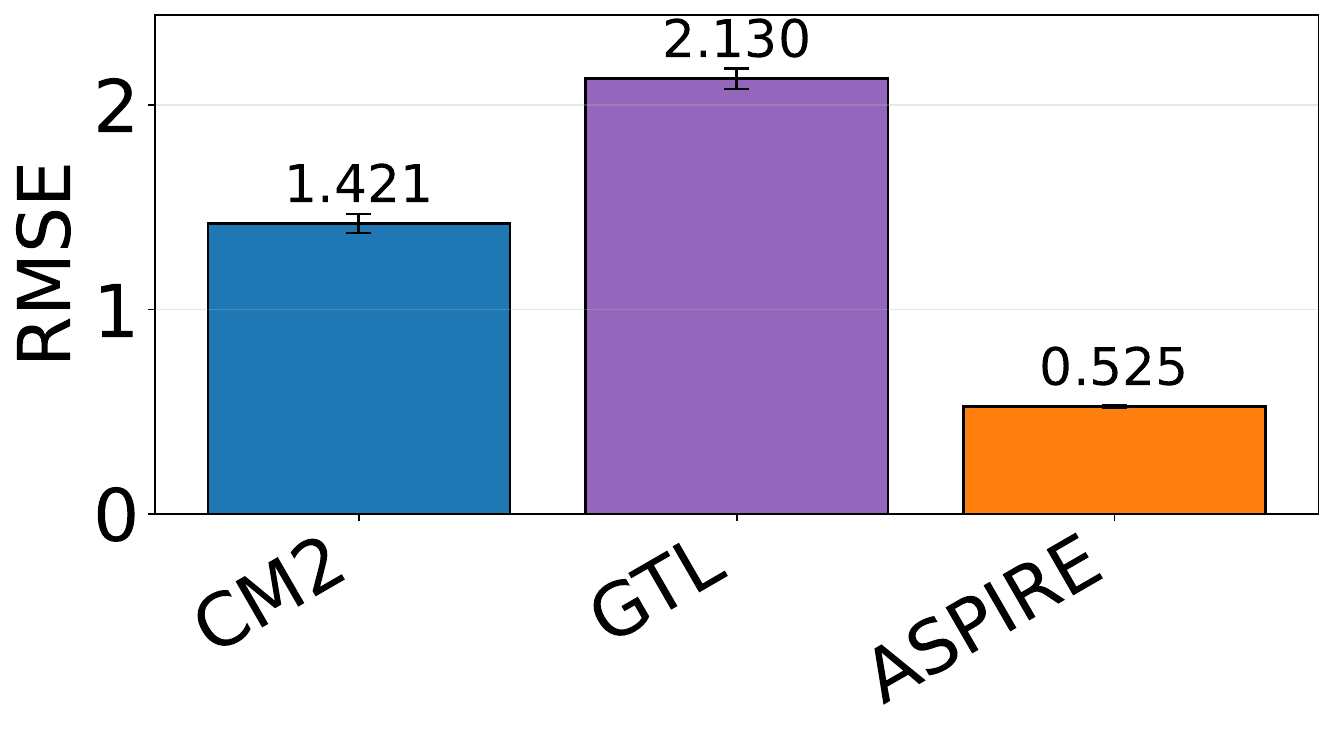}\vspace{-6pt}
    \caption{Zero-shot RMSE.}
    \label{fig:reg_0}
    \end{subfigure}
    \begin{subfigure}{\linewidth}
    \centering
    \includegraphics[width=\linewidth]{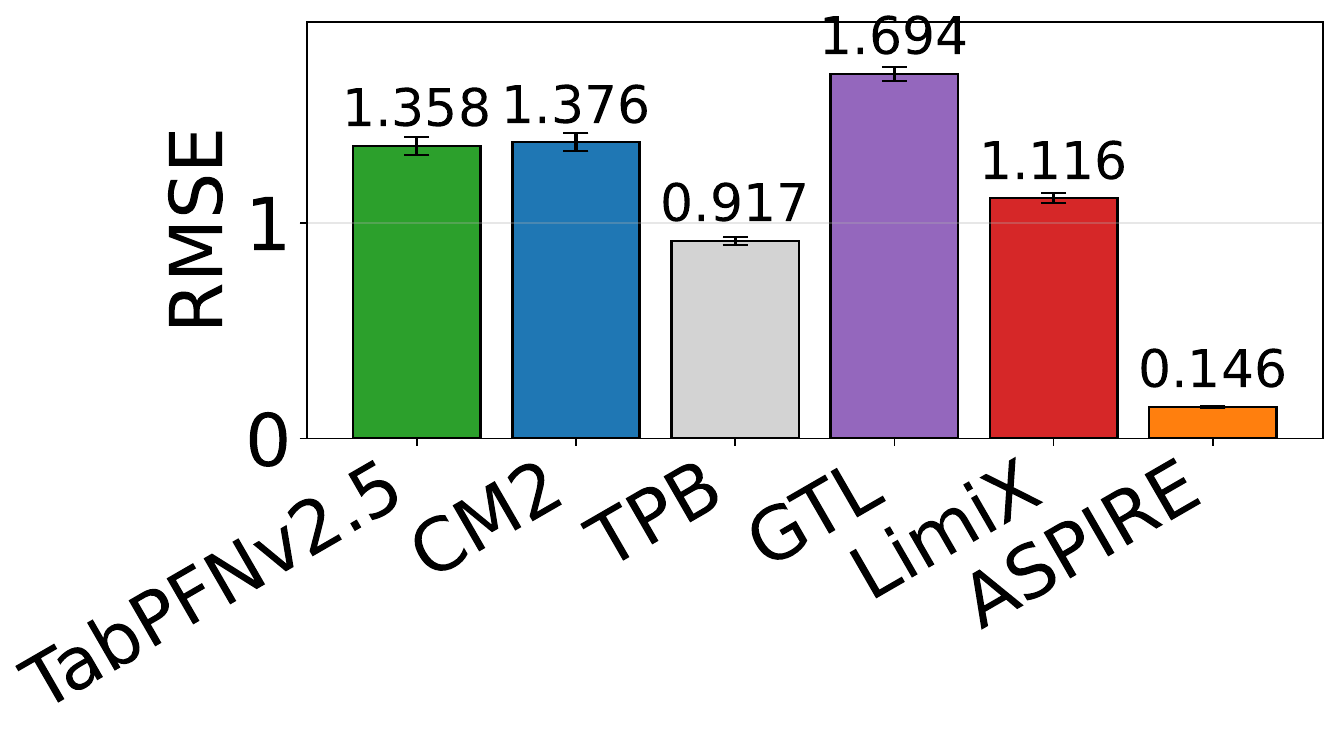}\vspace{-8pt}
    \caption{Five-Shot RMSE.}
    \label{fig:reg_5shot}
    \end{subfigure}
    \caption{Few-shot  average RMSE ($\downarrow$) across held out regression datasets.}
    
    \vspace{-12pt}
\end{wrapfigure}

We observe a similar trend in regression. As shown in Figures~\ref{fig:reg_0} and~\ref{fig:reg_5shot}, ASPIRE outperforms all baselines in both zero-shot and five-shot settings. Open-world few-shot regression has received comparatively little attention, and existing methods often rely on tokenized scalar targets 
(e.g., GTL~\cite{wen2024supervised}) or produce only point estimates (e.g., CM2~\cite{ye2024towards}), which empirically limits their performance. In contrast, the UNLI framework enables ASPIRE to directly model conditional probability densities over continuous targets, allowing it to more effectively leverage limited support sets and generalize to held-out regression tasks (full results in App.~\ref{sec:appendix}).




\subsubsection{Few-Shot Open-world AFA}\label{sec:afa}
\begin{wrapfigure}{r}{0.5\linewidth}
    \vspace{-12pt}
    \centering
    \begin{subfigure}{\linewidth}
    \centering
    \includegraphics[width=\linewidth]{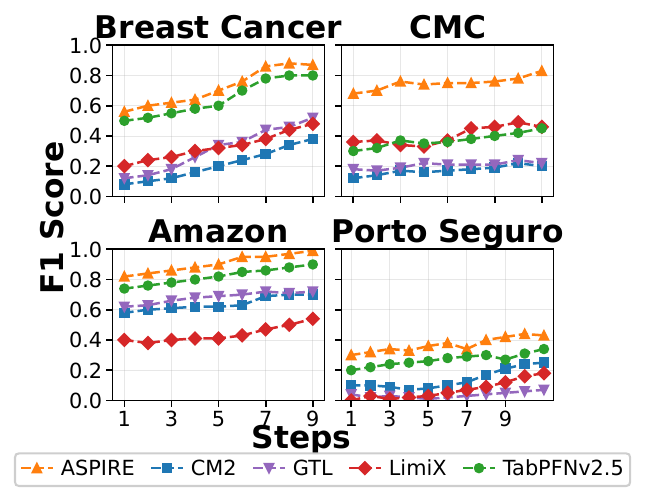}
    \caption{Active feature acquisition performance (F1 scores).}
    \label{fig:afa_fewshot}
    \end{subfigure}
    \begin{subfigure}{\linewidth}
    \centering
    \includegraphics[width=\linewidth]{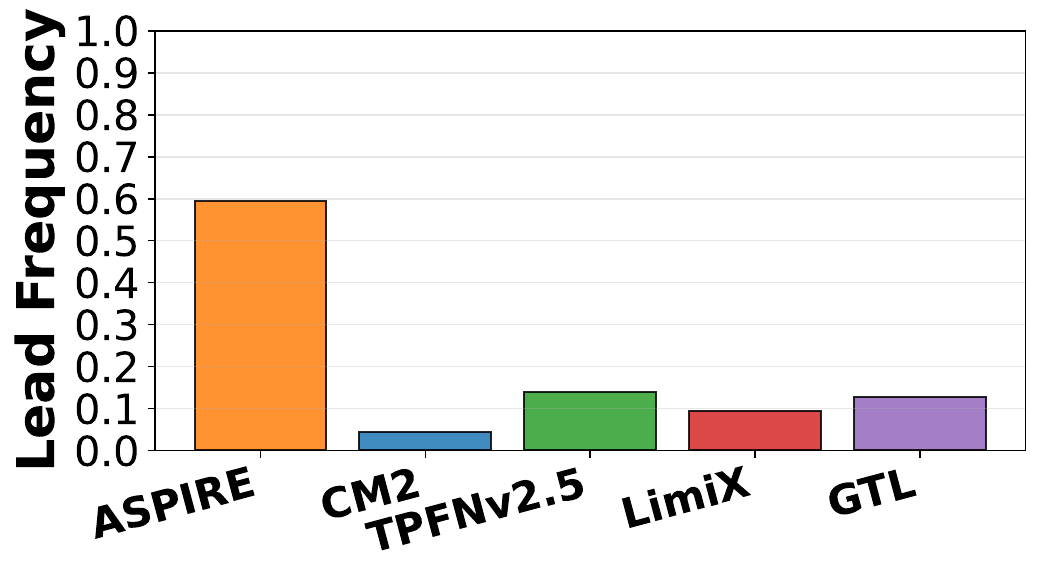}
    \caption{The \% of dataset steps each method leads the acquisition process averaged across datasets}
    \label{fig:afa_summary}
    \end{subfigure}
    \caption{Few-shot open-world AFA results.}
    
    \vspace{-12pt}
\end{wrapfigure}
We next evaluate models’ ability to assess feature informativeness in novel settings using the open-world active feature acquisition (AFA) framework described in Sec.~\ref{sec:afa_method}. In this setting, instances are partially observed at inference time and feature values are acquired sequentially at a cost. All methods employ an expected information gain (EIG) policy (Eq.~\ref{eq:eig}) to select features, with conditional distributions estimated by the corresponding model. Crucially, this requires reasoning over partially observed instances involving feature (and combinations) not seen during training. Models are provided a small support set (5-shot) to guide per-instance acquisition decisions, starting from an empty feature set.

Figure~\ref{fig:afa_fewshot} reports F1 performance as a function of the number of acquired features across unseen datasets. ASPIRE consistently achieves higher predictive performance with fewer acquisitions, indicating that its likelihood estimates enable more informative early feature selection. To summarize acquisition efficiency, Figure~\ref{fig:afa_summary} summarizes AFA across the 15 held-out classification datasets and shows the average percentage of steps that each method led. We see that ASPIRE leads a majority of the time and is leading $4.5\times$ more frequently than the next competing method. Additional results are provided in App.~\ref{sec:appendix}.

Overall, these results show that UNLI-enabled models can effectively operate in the proposed open-world AFA setting, and that ASPIRE’s design supports reasoning not only about target predictions, but also about feature importance.

\subsection{Many-Shot Transfer Learning}
\begin{figure}[ht]
    \centering
    \begin{subfigure}{.475\linewidth}
    \centering
    \includegraphics[width=\linewidth]{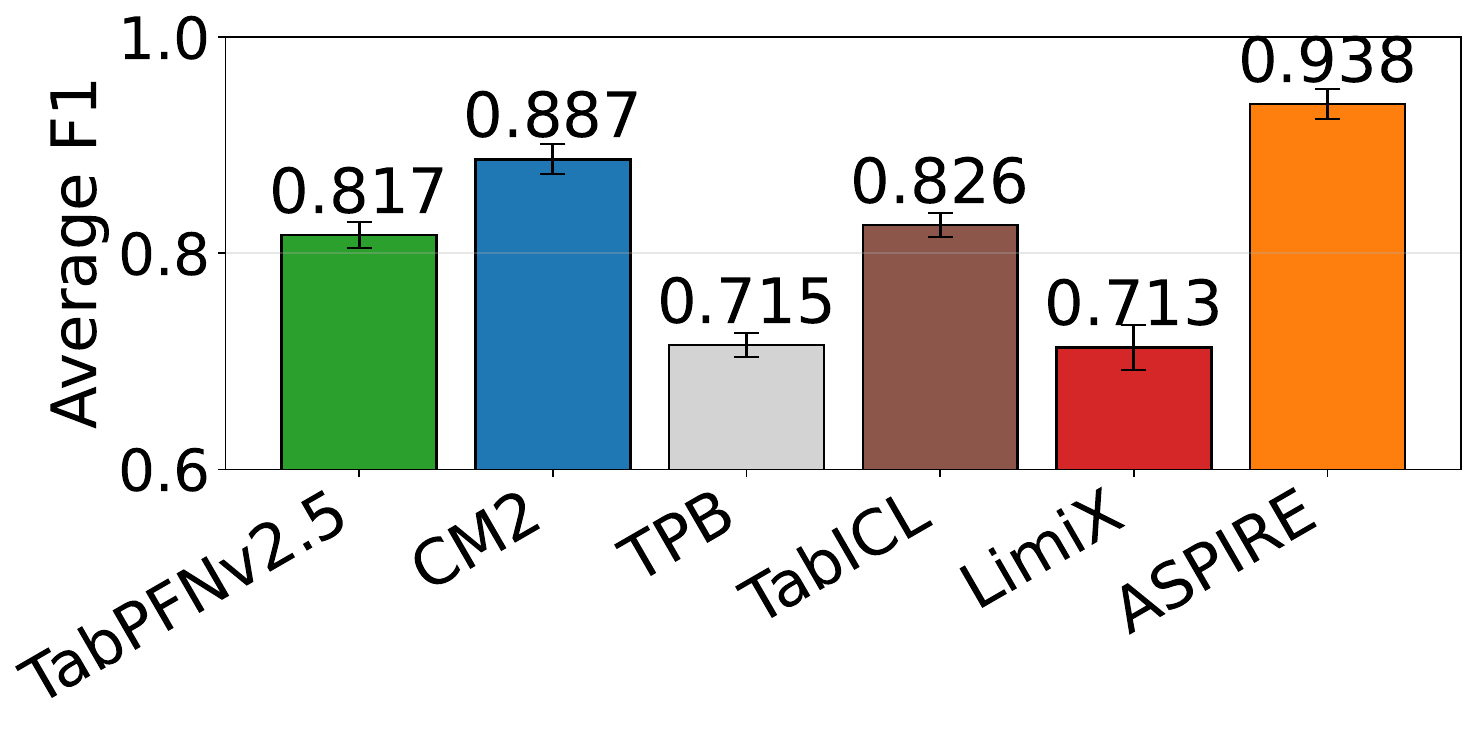}
    \caption{Classification F1 ($\uparrow$)}
    \label{fig:fine_class}
    \end{subfigure}
    \begin{subfigure}{.475\linewidth}
    \centering
    \includegraphics[width=\linewidth]{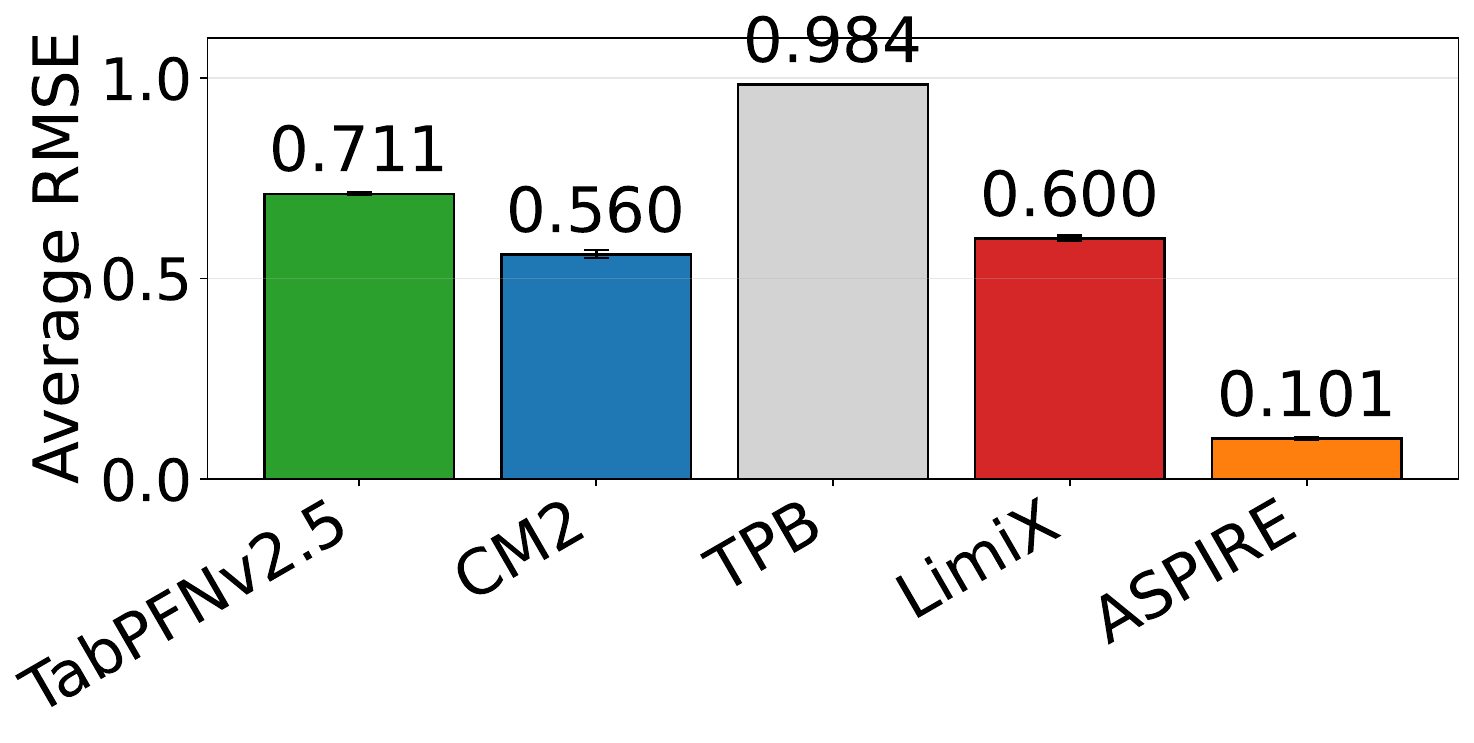}
    \caption{Regression RMSE ($\downarrow$)}
    \label{fig:fine_reg}
    \end{subfigure}
    \caption{Many-shot finetuning results across heldout datasets.}
    
    \vspace{-0pt}
\end{figure}
    
We next evaluate models’ ability to transfer under supervised adaptation when ample training data are available. Figures~\ref{fig:fine_class} and~\ref{fig:fine_reg} report many-shot transfer results for classification and regression, respectively. Each model is fine-tuned on the target dataset’s training split using AdamW, with early stopping on a held-out validation set, and evaluated on a fixed test split under a consistent 70\%/15\%/15\% train/validation/test protocol. In-context–learning-based models are fine-tuned following the procedure of \citet{hollmann2025accurate}; full implementation details are provided in App.~\ref{sec:appendix}.
Across datasets, ASPIRE achieves the strongest performance, indicating that its learned representations provide an effective initialization for supervised transfer and full-data adaptation. Taken together, these results underscore ASPIRE’s versatility: the same model supports both in-context few-shot inference—consistent with UNLI objectives—and end-to-end supervised fine-tuning.

\subsection{Ablation Studies} 

We systematically evaluate ASPIRE's key components, with the following key observations. (1) Semantic grounding using dataset context (e.g., as GTL \cite{wen2024supervised} does) is important. We see that removing this drops performance ($0.735$ $\rightarrow$ $0.696$ F1). Notwithstanding, even without this semantic grounding, ASPIRE is leading other methods in five shot learning tasks (even against methods 
like GTL \emph{with} semantic grounding), further showing the strength of our framework. (2) Permutation equivariant set-to-set mappings are important. We ablate the intra-instance set-to-set mappings (Eq.~\ref{eq:rho}), and instead directly use the atom-embeddings (\S\ref{sec:atoms}) directly passed to the aggregator (\S\ref{sec:aggregation}), which 
\begin{wraptable}{r}{0.22\textwidth}

\scriptsize
\setlength{\tabcolsep}{3pt}
\caption{ASPIRE Ablation. 5-shot learning F1 and RMSE averaged across datasets.}
\label{tab:aspire_ablation}
\begin{center}
\begin{tabular}{l c c}
\toprule
\textbf{Variant} & \textbf{F1 $\uparrow$} & \textbf{RMSE $\downarrow$} \\
\midrule

Full \textsc{ASPIRE} & 0.735 & 0.146 \\

\midrule
\multicolumn{3}{l}{\textbf{Semantic Grounding}} \\
\cmidrule(lr){1-3}
w/o Dataset Desc. & 0.696 & 0.183 \\

\midrule
\multicolumn{3}{l}{\textbf{Permutation Equivariance}} \\
\cmidrule(lr){1-3}
w/o IntraSet2Set & 0.690 & 0.215 \\

\midrule
\multicolumn{3}{l}{\textbf{Permutation Invariance}} \\
\cmidrule(lr){1-3}
Positional Enc. & 0.643 & 0.188 \\

\bottomrule
\end{tabular}
\end{center}
\end{wraptable}
drops performance ($0.735$ $\rightarrow$ $0.690$  F1), providing evidence that our permutation equivariant intra-instance design provides richer late fusion.
(3) Permutation invariance is important. We ablate our proposed type embeddings and instead utilize traditional positional encoding on all aggregation tokens (similar to a traditional LLM), which breaks permutation invariance at aggregation (\S\ref{sec:aggregation}), resulting in worse performance ($0.735\rightarrow0.643$ F1).

\section{Discussion} 
Prior work such as GTL~\cite{wen2024supervised} and CM2~\cite{ye2024towards} comes closest to the goals of universal neural likelihood inference (UNLI) by incorporating semantic grounding under limited supervision, but each remains fundamentally constrained: prompt-based approaches discretize scalar targets and are sensitive to prompt structure, while CM2's attention-based approach lacks in-context learning, late-fusion equivariance, and likelihood-based prediction. ASPIRE overcomes these limitations through a cohesive UNLI objective and set-based architecture, enabling principled likelihood estimation and strong performance across zero-, few-, and many-shot regimes. Our results further demonstrate that semantic grounding remains beneficial even with abundant labeled data, allowing ASPIRE to outperform large-scale in-context learning models that omit feature and dataset semantics—an important consideration given that most real-world datasets are semantically annotated. Moreover, by introducing open-world active feature acquisition in a cross-dataset setting, this work underscores the importance of calibrated uncertainty and arbitrary conditioning for sequential decision-making.

\textbf{In conclusion}, these findings suggest that semantic grounding, permutation-aware structure, and likelihood-based inference should be treated as first-class design principles for structured-data models. Together, they enable coherent reasoning across heterogeneous datasets, supervision regimes, and acquisition settings, and point toward a unified framework for open-world inference that integrates uncertainty, meaning, and data acquisition under realistic constraints.

\section{Impact Statement}
This paper presents work whose goal is to advance the field of machine learning. There are many potential societal consequences of our work, none of which we feel must be specifically highlighted here.

\bibliography{example_paper}
\bibliographystyle{icml/icml2026}

\newpage
\appendix
\onecolumn
\section{Appendix}
\label{sec:appendix}
\section{Additional Related Works}

\subsection{Set Modeling}
Early methods enforce permutation invariance by augmenting training data with randomly permuted versions of input sets, treating them as sequences while training models to produce consistent outputs across permutations. However, this approach does not guarantee invariance in practice—especially with finite data and limited model capacity—since sequence models inherently exploit positional information~\citep{zaheer2017deep}.

A foundational work in this area is DeepSets~\citep{zaheer2017deep}, which proves that any continuous permutation invariant function can be expressed as $f(S) = h\left(\sum_{x \in S} g(x)\right)$, where $g$ maps individual elements and $h$ aggregates the result. This leads to a simple yet expressive two-stage neural architecture. DeepSets also introduces permutation equivariant layers through shared transformations and pooling operations to capture intra-set dependencies. However, later work~\citep{wagstaff2019limitations} shows that the latent dimension of such architectures must grow at least linearly with the set size to maintain universal approximability, which may limit their practicality.

To model richer interactions between set elements, Set Transformer~\citep{lee2019set} replaces pooling with self-attention mechanisms. Attention layers are inherently permutation equivariant as they compute weighted sums over all elements. By combining these with attention-based pooling, Set Transformer yields permutation-invariant representations while capturing complex intra-set relationships—an approach that lays the foundation to our ASPIRE architecture. Several extensions have emerged to address specific limitations. Hölder-based power means, and quasi-arithmetic pooling strategies~\citep{kimura2024permutation} generalize sum and max pooling for increased expressivity. Other approaches like subset-invariant regularization~\citep{cohen2020regularizing} enforce permutation symmetry via learning objectives rather than architectural constraints. Recent work~\citep{wang2023polynomial} provides refined insights into the trade-offs between model width, depth, and set size for maintaining expressivity.

\subsection{Active Feature Acquisition}
Active Feature Acquisition (AFA) aims to selectively acquire informative features under budget constraints, rather than passively predicting a target from fully observed data. Classical approaches use cost-sensitive classifiers—such as decision trees~\citep{ling2004decision}, naive Bayes~\citep{chai2004test}, and margin-based learners~\citep{nan2014fast}—to jointly minimize prediction error and acquisition cost. More recent works frame AFA as a sequential decision-making process and propose various acquisition policies, including greedy information gain~\citep{ma2018eddi,gong2019icebreaker,li2020dynamic}, tree search~\citep{zubek2004pruning}, imitation learning~\citep{he2012imitation,he2016active}, and reinforcement learning~\citep{ruckstiess2011sequential,shim2018joint,zannone2019odin,li2021active,li2024distribution,li2022generative}. However, these approaches are domain-specific and require retraining when applied to new datasets. In contrast, our ASPIRE model enables AFA in an open-world setting—allowing feature acquisition on entirely novel datasets and domains without additional training.

\section{Data Processing}

\subsection{Training Data}
We use datasets from OpenTabs \citep{ye2024towards} for training, validation, and testing. We manually collect dataset descriptions and feature descriptions from UCI ML repository, Kaggle, OpenML, etc. Further, we curate metadata about the dataset by obtaining statistics about the dataset using python functions, for example, collecting potential classes for each target, data type of the feature values etc. Tables which have too few rows and columns, which have unclear and invalid data, are dropped. We will open-source this metadata upon publication. We identify task as classification or regression based on target value types, followed by min-max normalization of continuous values. Feature descriptions are embedded using pre-trained transformer models (BERT-base-uncased) to create dense semantic representations that capture feature semantics across diverse domains.

\subsection{Evaluation Data}
The following table ~\ref{tab:downstream} details information about the downstream datasets we use as test datasets in our evaluation.
\begin{table}[t]
\caption{Table of the downstream datasets in our experiments, along with different information}
\label{tab:downstream}
\begin{center}
\adjustbox{width=\columnwidth,center}%
{%
\scriptsize
\begin{tabular}{|l|c|r|r|r|r|l|}
\hline
Dataset Name & R/C & Samples & Numerical & Categorical & Label Classes & Source \\
\hline
Breast & C & 699 & 9 & 0 & 2 & https://archive.ics.uci.edu/dataset/15/breast+cancer+wisconsin+original \\
Bone & C & 1479 & 2 & 7 & 3 & https://archive.ics.uci.edu/dataset/3/connectionist+bench+choice \\
Diabetes & C & 768 & 8 & 0 & 2 & https://openml.org/d/37 \\
Vehicle & C & 846 & 18 & 0 & 4 & https://archive.ics.uci.edu/dataset/149/statlog+vehicle+silhouettes \\
Satimage & C & 6430 & 36 & 0 & 6 & https://archive.ics.uci.edu/dataset/146/statlog+landsat+satellite \\
Sick & C & 3772 & 7 & 22 & 2 & http://archive.ics.uci.edu/dataset/102/thyroid+disease \\
Analcatdata & C & 797 & 0 & 4 & 6 & https://pages.stern.nyu.edu/jsimonof/AnalCatData/Data/ \\
Pcl & C & 1109 & 21 & 0 & 2 & https://openml.org/d/1068 \\
Adult & C & 48842 & 6 & 8 & 2 & https://archive.ics.uci.edu/dataset/2/adult \\
PhishingWebsites & C & 11055 & 0 & 30 & 2 & https://archive.ics.uci.edu/dataset/327/phishing+websites \\
Cylinder-bands & C & 540 & 18 & 21 & 2 & https://archive.ics.uci.edu/dataset/32/cylinder+bands \\
MiceProtein & C & 1080 & 77 & 4 & 8 & https://archive.ics.uci.edu/dataset/342/mice+protein+expression \\
Car & C & 1728 & 0 & 6 & 4 & https://archive.ics.uci.edu/dataset/19/car+evaluation \\
Segment & C & 2310 & 19 & 0 & 7 & http://archive.ics.uci.edu/dataset/50/image+segmentation \\
Porto-seguro & R & 2000 & 26 & 31 & 2 & https://openml.org/d/44787 \\
Amazon & C & 2000 & 0 & 9 & 2 & https://openml.org/d/44712 \\
Elevators & R & 16599 & 18 & 19 & - & https://openml.org/d/216 \\
Yprop & R & 8885 & 251 & 0 & - & https://openml.org/d/416 \\
Topo & R & 8885 & 266 & 267 & - & https://openml.org/d/422 \\
SAT11 & R & 4400 & 115 & 1 & - & https://www.cs.ubc.ca/labs/algorithms/Projects/SATzilla/ \\
Diamonds & R & 53940 & 6 & 3 & - & https://openml.org/d/42225 \\
House\_sales & R & 21613 & 20 & 1 & - & https://openml.org/d/42731 \\
\hline
\end{tabular}
}
\end{center}
\end{table}


\section{ASPIRE Architecture}\label{sec:appendix_method}
\begin{figure}[ht]
\centering
\includegraphics[width=\linewidth,trim={0 0 0 3cm},clip]{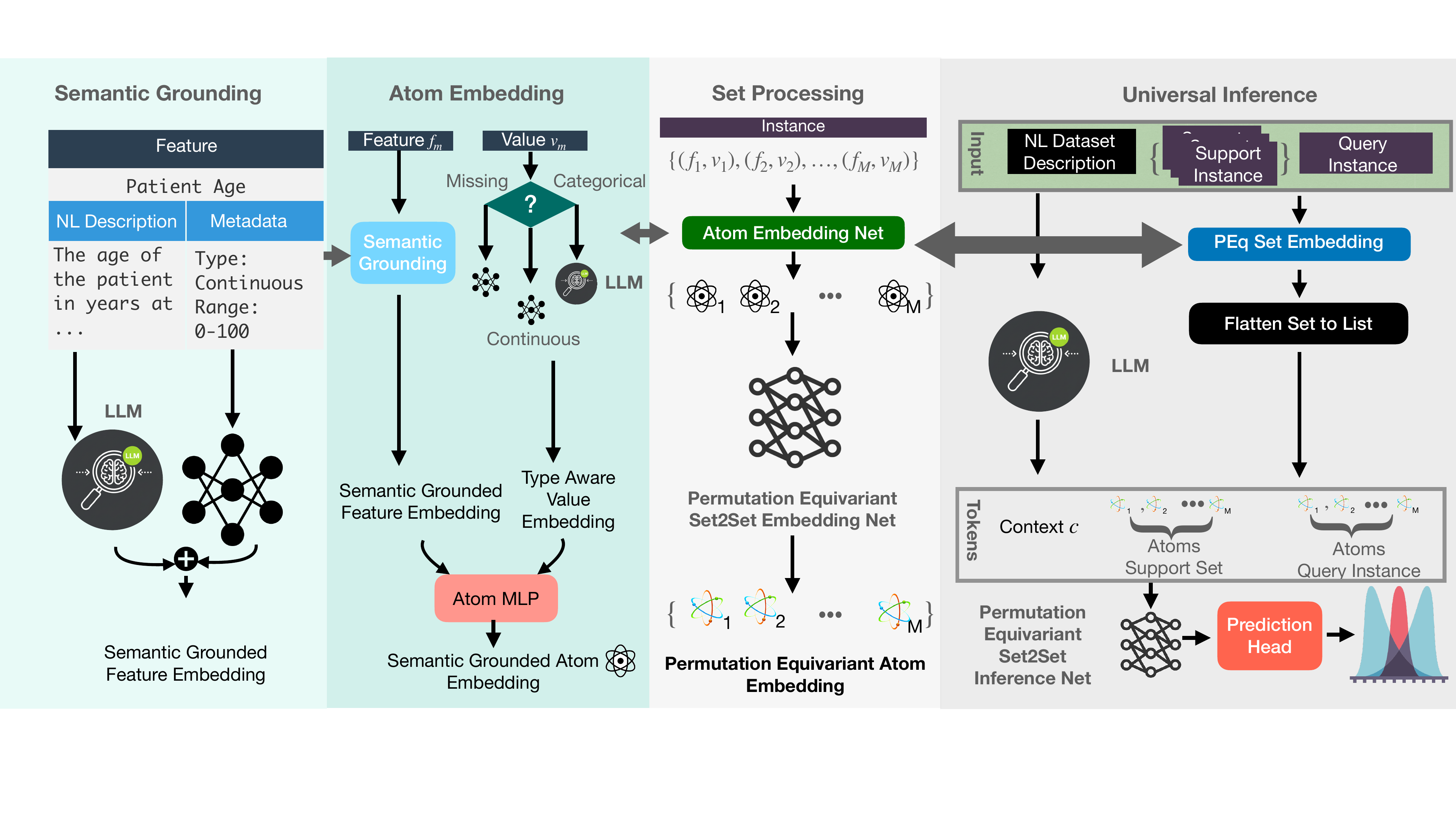}
\vspace{-38pt}
\caption{ASPIRE processing pipeline. The framework transforms heterogeneous inputs from different domains through semantic alignment into a shared understanding space, applies permutation-invariant set processing to capture feature interactions while maintaining order independence, and produces universal predictions that support arbitrary conditioning and cross-domain transfer.}
\label{fig:highlevel}
\end{figure}

In addition to ASPIRE model descriptions in the paper, the following implementation details are relevant. Our optimization uses AdamW with a learning rate of 1e-4 and cosine annealing with warm restarts (100 warmup steps, weight decay of 0.01-0.04). During training, we apply feature dropout with a 40\% masking rate, where selected features have their values replaced with missing value embeddings to improve generalization across diverse tabular domains.
 Feature encoding uses specialized approaches: real-valued features utilize Fourier embeddings with 256 learnable frequency components spanning 1-256, while categorical features undergo BERT encoding followed by projection to the model dimension of 768. The Set Transformer components use 4-8 attention heads with 16-32 induced points for permutation-invariant row representations.

The multi-task learning framework combines cross-entropy loss for classification and mixture-of-Gaussians negative log-likelihood for regression, where the regression head employs 10 Gaussian components with learnable weights, means, and log-variances. Training incorporates early stopping with patience of 5-7 epochs. High level pipeline is shown in Fig. ~\ref{fig:highlevel}

\section{Additional Results}

\label{sec:appx_additional}
\textbf{CM2}: We use the pre-trained model, which have number of transformer layers 3, attnetion heads 8, batch size 256, learning rate for finetuning 3e-4, patience is 5. 
For performing few-shot learning, \citet{ye2024towards} prescribes $k$-shot learning via finetuning the model with $k$ examples sampled from the train set. We observe that this causes high-variance in the performance, therefore we report the mean of all metrics for CM2 few-shot learning experiments.

\textbf{GTL}: We use the authors’ released GTL model (a LLaMA-2 backbone; 7B/13B parameters depending on the checkpoint) and the official inference pipeline. Unless otherwise stated, we keep the default evaluation settings provided by the authors and do not perform dataset-specific hyperparameter tuning. When we run $k$-shot adaptation for GTL, we follow the training configuration reported in the GTL work (AdamW optimizer, learning rate $1\times10^{-5}$, batch size $512$, max sequence length $4096$, and $256$ update steps).

\textbf{LimiX}: We evaluate the publicly released LimiX checkpoints (LimiX-16M, and LimiX-2M when reporting the lightweight variant). LimiX is used in its training-free, context-driven inference mode: we provide the labeled training split as context and query the model for predictions on the test split, without any dataset-specific fine-tuning. We run the official predictor/inference interface and use the default inference configuration from the release (enabling mixed precision when available).

\textbf{TabICL}: We use the pre-trained TabICL checkpoint and the official inference code. TabICL performs dataset-wise in-context learning by conditioning on the labeled training set and predicting the test set in a single forward pass without parameter updates. For large tables, we retain the authors’ default memory-management behavior, including automatic batch-size selection based on available GPU memory and optional CPU/disk offloading.

\textbf{TabPFN v2.5}: We use the official TabPFN-2.5 release via the authors’ provided implementation, and evaluate it in the standard PFN in-context regime (training split as context; no parameter updates). We do not tune hyperparameters per dataset and reuse the library-default settings for prediction.

\textbf{TP-BERTa}: We add TP-BERTa as an additional LM-based tabular baseline. TP-BERTa is initialized from a RoBERTa backbone adapted for tabular prediction and is designed for standard supervised tabular prediction via fine-tuning on downstream datasets. Following the authors’ guidance, we fine-tune TP-BERTa per dataset with a larger training budget (up to 200 epochs) and early stopping with patience 50 epochs.
\textbf{FeatLLM}: We follow FeatLLM’s feature-engineering pipeline, which uses an LLM to propose candidate feature transformations and trains a downstream linear model on the resulting engineered features. We generate $20$ feature ensembles with $10$ rules per ensemble, train the downstream model using Adam with learning rate $0.01$ for $200$ epochs, and report the cross-validation performance as prescribed by the FeatLLM setup.

\textbf{TabLLM}: We implement TabLLM by serializing each row into natural language using PromptSource-style templates and verbalizers, and we fine-tune a sequence-to-sequence LLM using the T-Few recipe. Following the TabLLM setup, we use a T0 backbone (TabLLM uses T0-11B in the original work; in compute-limited settings we use a smaller T0 variant) with a maximum input length of $1024$ tokens, and we follow the default TabLLM/T-Few hyperparameters, including training for $30$ epochs on public datasets. For stability in few-shot settings, we report averages across random draws/seeds where applicable.

\begin{table}[t]
\centering
\caption{\textbf{Replication check for TabLLM (4-shot, AUC).} We report self-run AUC on our local implementation alongside paper-reported AUC and $\Delta$ (self-run $-$ paper). We use the T0-3B backbone due to hardware constraints. Values are in percentage points; higher is better.}
\label{tab:tabllm_replication_auc_4shot}
\scriptsize
\setlength{\tabcolsep}{5pt}
\begin{tabular}{lccc}
\toprule
\textbf{Dataset} & \textbf{Self-run AUC} & \textbf{Paper AUC} & $\boldsymbol{\Delta}$ \\
\midrule
Bank        & 49.69 & 60.00 & -10.31 \\
Blood       & 41.58 & 47.00 & -5.42 \\
Calhousing  & 58.81 & 59.00 & -0.19 \\
Car         & 78.17 & 80.00 & -1.83 \\
Credit-g    & 56.02 & 65.00 & -8.98 \\
Diabetes    & 56.58 & 57.00 & -0.42 \\
Heart       & 63.16 & 68.00 & -4.84 \\
Income      & 72.80 & 77.00 & -4.20 \\
Jungle      & 55.53 & 63.00 & -7.47 \\
\midrule
\textbf{Average} & \textbf{59.15} & \textbf{64.00} & \textbf{-4.85} \\
\bottomrule
\end{tabular}
\end{table}
\begin{table}[ht]
\centering
\caption{\textbf{Replication check for FeatLLM (4-shot, AUC).} We report self-run AUC on our local implementation alongside paper-reported AUC and the difference $\Delta$ (self-run $-$ paper). Values are in percentage points; higher is better.}
\label{tab:featllm_replication_auc}
\scriptsize
\setlength{\tabcolsep}{5pt}
\begin{tabular}{lccc}
\toprule
\textbf{Dataset} & \textbf{Self-run AUC} & \textbf{Paper AUC} & $\boldsymbol{\Delta}$ \\
\midrule
adult        & 84.85 & 86.68 & -1.83 \\
bank         & 69.92 & 70.45 & -0.53 \\
blood        & 65.22 & 68.34 & -3.12 \\
car          & 61.02 & 72.69 & -11.67 \\
communities  & 63.74 & 75.39 & -11.65 \\
credit-g     & 47.56 & 55.94 & -8.38 \\
diabetes     & 75.50 & 80.28 & -4.78 \\
heart        & 82.08 & 75.66 & 6.42 \\
\midrule
\textbf{Average} & \textbf{68.74} & \textbf{73.18} & \textbf{-4.44} \\
\bottomrule
\end{tabular}
\end{table}
\begin{table}[ht]
\centering
\caption{\textbf{Replication check for LimiX-16M on the BCCO benchmark.}
We reproduce the BCCO classification (AUC/ACC/F1; higher is better) and regression (RMSE; lower is better, and $R^2$; higher is better) results for LimiX-16M.
We report our self-run averages, the paper-reported averages, and $\Delta$ (self-run $-$ paper).}
\label{tab:limix_replication_bcco}
\scriptsize
\setlength{\tabcolsep}{3.5pt}
\begin{tabular}{l|ccc|ccc|ccc}
\toprule
& \multicolumn{3}{c|}{\textbf{AUC} $\uparrow$}
& \multicolumn{3}{c|}{\textbf{ACC} $\uparrow$}
& \multicolumn{3}{c}{\textbf{F1} $\uparrow$} \\
\cmidrule(lr){2-4} \cmidrule(lr){5-7} \cmidrule(lr){8-10}
\textbf{Setting} 
& \textbf{Self} & \textbf{Paper} & $\boldsymbol{\Delta}$
& \textbf{Self} & \textbf{Paper} & $\boldsymbol{\Delta}$
& \textbf{Self} & \textbf{Paper} & $\boldsymbol{\Delta}$ \\
\midrule
BCCO-CLS & 86.02 & 87.10 & -1.08
         & 78.60  & 80.40 & -1.79
         & 69.77 & 73.10 & -3.33 \\
\midrule
& \multicolumn{3}{c|}{\textbf{RMSE} $\downarrow$}
& \multicolumn{3}{c|}{\textbf{$R^2$} $\uparrow$}
& \multicolumn{3}{c}{ } \\
\cmidrule(lr){2-4} \cmidrule(lr){5-7}
\textbf{Setting}
& \textbf{Self} & \textbf{Paper} & $\boldsymbol{\Delta}$
& \textbf{Self} & \textbf{Paper} & $\boldsymbol{\Delta}$
& \multicolumn{3}{c}{ } \\
\midrule
BCCO-REG & 38.78 & 38.60 & 0.18
         & 78.99 & 79.40 & -0.41
         & \multicolumn{3}{c}{ } \\
\bottomrule
\end{tabular}
\end{table}
\begin{table}[t]
\centering
\caption{\textbf{Replication check for GTL on unseen Kaggle holdout tasks.}
We report self-run performance on the GTL holdout split alongside paper-reported values and $\Delta$ (self-run $-$ paper).
Results are averaged over 20 classification and 24 regression tasks.}
\label{tab:gtl_replication_holdout}
\scriptsize
\setlength{\tabcolsep}{6pt}
\begin{tabular}{lccc}
\toprule
\textbf{Setting} & \textbf{Self-run} & \textbf{Paper} & $\boldsymbol{\Delta}$ \\
\midrule
Classification (avg.) & 0.684 & 0.714 & -0.030 \\
Regression (avg.)     & 0. 301 & 0.288 & - 0.013\\
\bottomrule
\end{tabular}
\end{table}

ASPIRE's arbitrary conditioning capability naturally extends to Active Feature Acquisition (AFA), where features are sequentially acquired to minimize cost while maximizing prediction accuracy. Unlike existing methods that require per-dataset training, ASPIRE performs AFA on novel datasets without retraining by leveraging its universal inference framework.

Following prior work~\citep{li2021active}, we formulate AFA as a sequential decision-making problem where features are acquired based on estimated mutual information with the target variable. ASPIRE's probabilistic predictions enable principled feature selection under budget constraints, while its cross-dataset knowledge transfer provides robust performance even on previously unseen domains. EDDI \cite{ma2018eddi} and DIME \cite{gadgilestimating} are classical information-based acquisition baselines for active feature acquisition (AFA). In our experiments, we follow prior work~\citep{li2021active} and instantiate both methods as sequential policies that, at each step, choose the next unobserved feature to acquire by scoring candidates with an information-theoretic utility computed under a probabilistic predictive model. Concretely, EDDI selects the feature that maximizes the expected information gain about the target, operationalized as the expected reduction in predictive uncertainty after observing that feature, while DIME similarly ranks features by their estimated mutual-information contribution but uses a decomposed/efficient approximation to make this computation tractable across many candidates. We apply EDDI and DIME in the same acquisition protocol and budgets as ASPIRE, using their respective MI-based scores to produce an ordered sequence of feature acquisitions, and report downstream predictive performance (e.g., F1) as a function of acquisition step to enable a direct, apples-to-apples comparison of acquisition quality.This demonstrates ASPIRE's flexibility beyond standard prediction tasks, supporting cost-aware inference in open-world settings. Detailed AFA algorithms and experimental results are provided below in this section.


The information reward for acquiring feature $i$ given observed features $\mathbf{x}_o$ is:
\begin{equation}
\begin{split}
R(i, \mathbf{x}_o) &= \mathbb{E}_{\mathbf{x}_i \sim p(\mathbf{x}_i|\mathbf{x}_o)} D_{KL}[q(\mathbf{z}|\mathbf{x}_i, \mathbf{x}_o) \| q(\mathbf{z}|\mathbf{x}_o)] \\
&\quad - \mathbb{E}_{\mathbf{x}_\phi, \mathbf{x}_i \sim p(\mathbf{x}_\phi, \mathbf{x}_i|\mathbf{x}_o)} \\
&\quad\quad D_{KL}[q(\mathbf{z}|\mathbf{x}_\phi, \mathbf{x}_i, \mathbf{x}_o) \| q(\mathbf{z}|\mathbf{x}_\phi, \mathbf{x}_o)]
\end{split}
\end{equation}
\begin{itemize}
\item $\mathbf{x}_o$: Currently observed features
\item $\mathbf{x}_i$: Candidate feature $i$ to be acquired
\item $\mathbf{x}_\phi$: Target variables (labels)
\item $q(\mathbf{z}|\cdot)$: Posterior encoder distribution in VAE
\item $p(\mathbf{x}_i|\mathbf{x}_o)$: Conditional distribution of feature $i$ given observed features
\item $D_{KL}[\cdot \| \cdot]$: Kullback-Leibler divergence
The following algorithm describes our greedy procedure to select the next candidate from the set of unobserved features. 
\end{itemize}




\begin{figure*}[t]
    \centering
    \includegraphics[width=\textwidth]{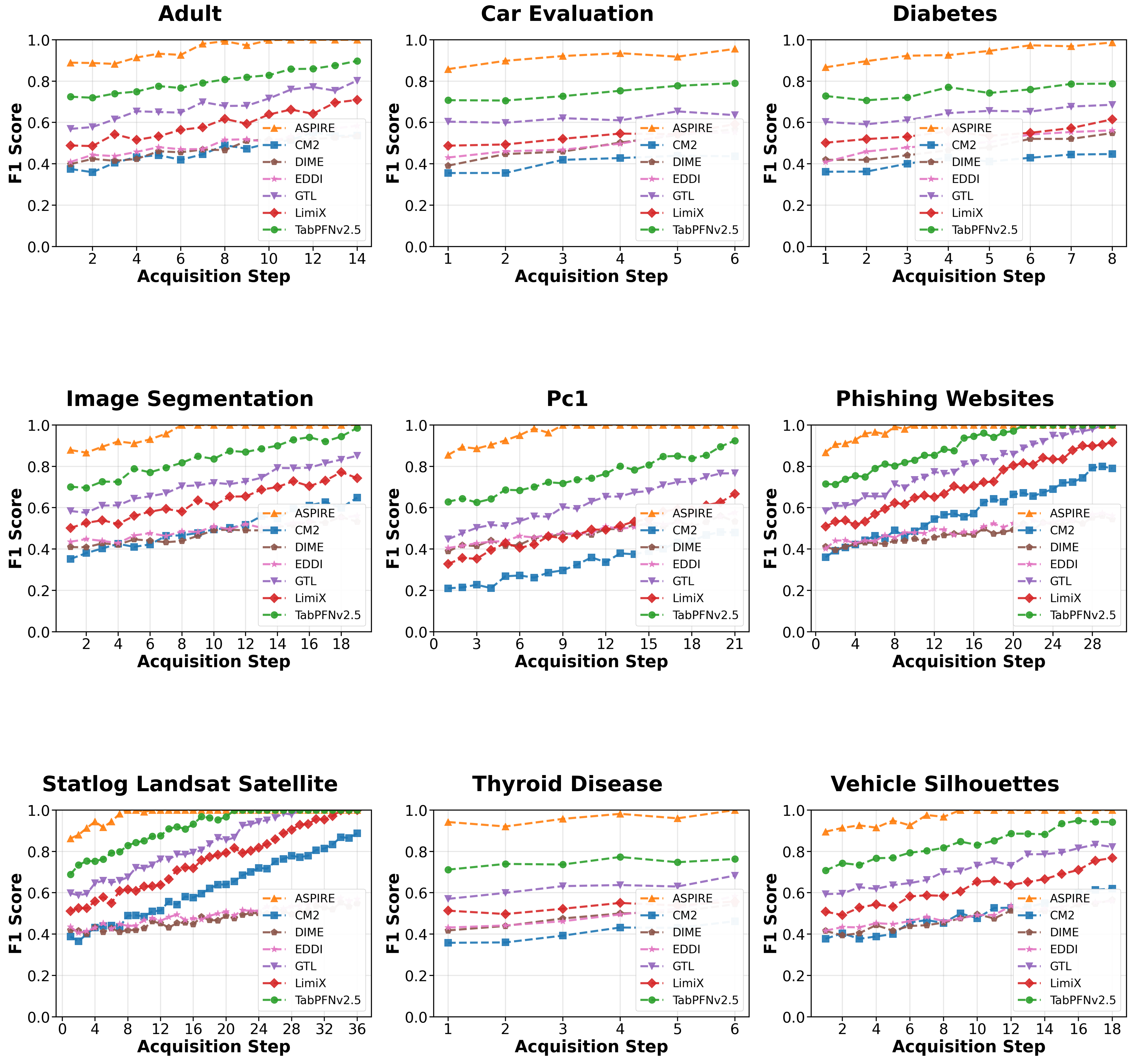}
    \caption{Active feature acquisition performance for Baselines finetuned on the task dataset (F1) across datasets as a function of acquisition step.}
    \label{fig:afa_multiplot}
\end{figure*}

\end{document}